\documentclass[letterpaper, 10 pt, conference]{ieeeconf}  

\usepackage{hyperref, bookmark}						
\hypersetup{colorlinks=true,linkcolor=black,citecolor=blue,urlcolor=blue}

\IEEEoverridecommandlockouts                              

\usepackage{cite}
\usepackage[flushleft]{threeparttable}
\usepackage{amsmath,amssymb,amsfonts}
\usepackage{algorithmic}
\usepackage{graphicx}
\usepackage{textcomp}
\usepackage{amsmath}

\usepackage{xcolor}
\usepackage{booktabs}
\usepackage{comment}
\usepackage{stackengine}

\usepackage{color}
\def\BibTeX{{\rm B\kern-.05em{\sc i\kern-.025em b}\kern-.08em
    T\kern-.1667em\lower.7ex\hbox{E}\kern-.125emX}}

\usepackage{hyperref, bookmark}						
\hypersetup{colorlinks=true,linkcolor=black,citecolor=blue,urlcolor=blue}

\title{\LARGE \bf
Open-Ended Multi-Modal Relational Reasoning for Video Question Answering}

\author{Haozheng Luo$^{1,3,4}$, Ruiyang Qin$^{2,4}$, Chenwei Xu$^{1}$, Guo Ye$^{1}$, and Zening Luo$^{1}$ 
\thanks{*This work was not supported by any organization}
\thanks{$^{1}$ Haozheng Luo, Chenwei Xu, Guo Ye and Zening Luo with Department of Electrical Engineering and Computer Science, Northwestern University, Evanston.
        {\tt\small \{robinluo2022, ChenweiXu2023, guoye2018, tonyluo2023\}@u.northwestern.edu }}%
\thanks{$^{2}$ Ruiyang Qin with Department of Computer Science and Engineering,
      University of Notre Dame, Notre Dame, IN
        {\tt\small rqin@nd.edu}}%
\thanks{$^{3}$ Haozheng Luo is the corresponding author.}%
\thanks{$^{4}$ These authors contributed equally to this work}%
}

\begin{document}

\maketitle
\thispagestyle{empty}
\pagestyle{empty}

\begin{abstract}
In this paper, we introduce a robotic agent specifically designed to analyze external environments and address participants' questions. The primary focus of this agent is to assist individuals using language-based interactions within video-based scenes. Our proposed method integrates video recognition technology and natural language processing models within the robotic agent. We investigate the crucial factors affecting human-robot interactions by examining pertinent issues arising between participants and robot agents. Methodologically, our experimental findings reveal a positive relationship between trust and interaction efficiency. Furthermore, our model demonstrates a 2\% to 3\% performance enhancement in comparison to other benchmark methods.


\end{abstract}
\vspace{-0.05in}
\section{Introduction}
\vspace{-0.05in}

We present a novel robotic agent specifically designed to assist visually impaired individuals by employing environmental perception and natural language interactions. In contrast to existing approaches \cite{9209377, 10.1145/3570731, 7549220, doi:10.5772/60416}, our agent offers advanced multi-modal relational reasoning capabilities and adaptive, personalized responses. Our technique incorporates the Relation-Aware Self-Attention mechanism \cite{wang2021ratsql} to address relational reasoning queries effectively. We implement this agent using the ROS system \cite{quigley2009ros}, Kobuki platform, and Python programming language. To evaluate our design, we initially establish a set of hypotheses and proceed to validate them through a series of experiments. Our findings reveal that our model outperforms competing methods, achieving a performance improvement of 2\% to 3\%.
Our robotic agent utilizes a multi-model approach in video question answering (VQA) process. Robotic agent involves a model based on the R(4+1)D  ~\cite{tran2018closer} with Resnet ~\cite{7780459} model to detect the relation and movement of the objects in the videos and put those scene information into the reasoning functions with the model of the BERT ~\cite{devlin2018bert} and XLNET ~\cite{yang2019xlnet}. We also employ the Relation-Aware Self-Attention mechanism for processing semi-structured input sequences and integrate our model with a video processing pipeline that includes ResNet-based object detection and optical flow analysis.  
\begin{figure}[!tb]
  \centering
  {\includegraphics[width=\linewidth,height=10cm]{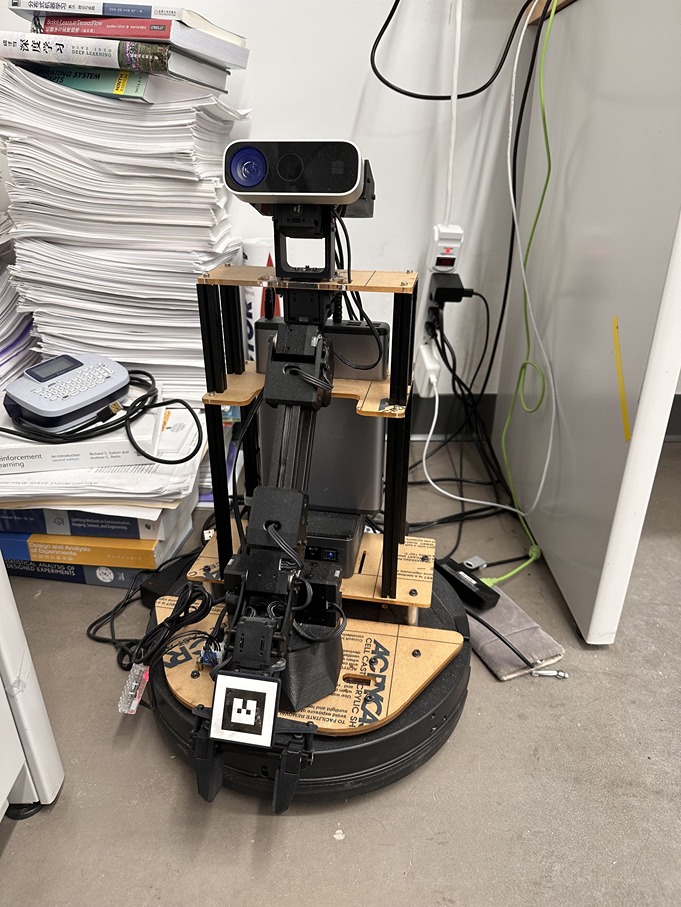}}
  \caption{\small The picture of the Kobuki Robot we use in paper}
  \label{figure:robot}
  \vspace{-0.25in}
\end{figure}
In our study, we initially formulated several hypotheses regarding the interaction between robots and users. Following this, we outlined the composition of our participant group and provided a description of the robot employed in our experiment. For each hypothesis, we designed a corresponding experiment to validate our assumptions. Lastly, we introduced the measurement techniques utilized for evaluating each hypothesis in our research.

We demonstrate superior performance by comparing our model performance and testing trust and interactivity of human robotics interaction.  When contrasted with other ViQA models, our approach exhibits a performance improvement of approximately 2\% to 3\%. To quantify trust and interactivity, we employ a survey-based methodology. The survey results indicate a positive correlation between trust and interactivity, supporting the superiority of our proposed model.

The rest of this paper starts by a summary of prior research on pre-trained language models and video recognition, followed by we provide an overview of our approach. Next, we introduce our implementation and the platform on which we executed our model. Furthermore, we elaborate on our experimental design and present our numerical results in detail. Lastly, we conclude with a discussion of potential future avenues for exploration.

\vspace{-0.05in}
\section{Related work}
\vspace{-0.05in}
This work has two related fields: {\sf pre-trained Language Model} and {\sf Video Recognition}.
\vspace{-0.05in}
\subsection{Pre-trained Language Model}
\vspace{-0.05in}
A variety of pre-trained Natural Language Networks have been developed for NLP applications. For example, the Neural-Symbolic Visual Question Answering (NS-VQA) model~\cite{yi2018neural}, employs a sequence-to-sequence approach for visual question answering (ViQA). This method is applicable in numerous everyday situations, such as dialogues. However, the approach described in VQA~\cite{yi2018neural} is unable to process linguistic context and comprehend grammar, both of which are essential for communication between a robot and a visually impaired individual \cite{luo2019question, hou2020fewshot, zhou2022learning,10463042}.

Previous research has focused on improving word embeddings \cite{mikolov2013distributed, peters2018deep, samel2021design}. With the groundbreaking transformation introduced by Vaswani et al. \cite{vaswani2017attention}, significant advancements have been made in the NLP domain, particularly regarding polysemous conditions in sentences. Recently, numerous pre-trained models have been developed to evaluate questions and generate answers, including BERT \cite{devlin2018bert}, XLNet \cite{yang2019xlnet}, InstructGPT \cite{NEURIPS2022_b1efde53}, and ViLBERT \cite{lu2019vilbert}. BERT is arguably the most popular among these models, owing to its simplicity and superior performance \cite{ qin2021ibert,liu2022sciannotate, 10.1093/bioinformatics/btab083}.

\begin{figure}
    \centering
    \includegraphics[width=\linewidth]{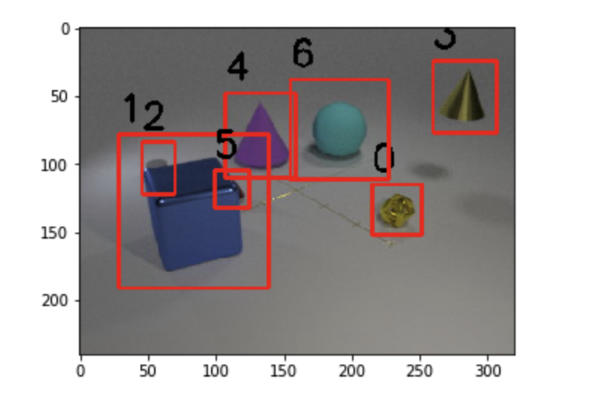}
    \caption{\small Example of bounding box detection in the CATER dataset}
    \label{fig:bbox}
    \vspace{-0.3in}

\end{figure}

\vspace{-0.05in}
\subsection{Video Recognition}
\vspace{-0.05in}
Numerous significant works have been conducted in addressing the CATER dataset ~\cite{girdhar2019cater}. The format of videos in the CATER dataset (as seen in fig \ref{fig:bbox}) makes it inefficient to use frame-wise scene parsing techniques, such as Mask-RNN ~\cite{he2017mask}. With the concept of 3D-ConvNet ~\cite{li20183d}, the groundbreaking video action recognition method called I3D ~\cite{carreira2017quo} emerged in the research field.

The I3D employs a two-stream 3D-ConvNet, where the inputs for the two streams are optical flow changes and image changes, respectively. The optical flow \cite{7410673} represents the motion field relationship. Building upon the I3D, we develop a model using I(2+1)D with ResNet \cite{he2015deep}, known as R(2+1)D.

There are several benefits in comparing the original 3D convolutions model with spatial and temporal features (like I3D) to the model that explicitly factorizes 3D convolution (like I(2+1)D). One advantage is the introduction of additional nonlinear rectification between the two operations. Another benefit is the explicit factorization of 3D convolution, making it easier to optimize compared to the original 3D convolutions model with spatial and temporal convolutions combined into a single operation. Despite R(2+1)D using only one kind of residual block uniformly, it still achieves state-of-the-art action recognition accuracy ~\cite{tran2018closer}.

\subsection{Foundation Model}
Additionally, the development of the transformer architecture has made transformer-based foundation models \cite{bommasani2021opportunities} significantly beneficial in classification tasks. They play a central role not only in machine learning but also in a wide range of scientific domains, such as ChatGPT \cite{brown2020language,floridi2020gpt} for natural language, BloombergGPT \cite{wu2023bloomberggpt} for finance, DNABERT \cite{zhou2024dnabert,zhou2023dnabert,ji2021dnabert} for genomics, and many others \cite{qin2021ibert,Luo_2023,liu2022sciannotate,luo2024decoupled,yu2024enhancing,zhang2024smutf}. The Modern Hatfield Network provides another efficient method to do the question and answering classfication with the memory retrieval.
Modern Hopfield models \cite{hu2024nonparametric,hu2024computational,wu2023stanhop,hu2023SparseHopfield,hopfeildblog2021,ramsauer2020hopfield,hu2024outlier} showcase fast convergence and exponential memory capacity, linking them to Transformer architecture as advanced attention mechanisms. Their application spans various domains, including drug discovery \cite{schimunek2023contextenriched}, immunology \cite{widrich2020modern}, and time series forecasting \cite{wu2023stanhop,hu2023SparseHopfield,auer2023conformal}, signifying their influence on future large-scale model designs. 


\begin{figure}[ht]
  \centering
  \includegraphics[width=\linewidth]{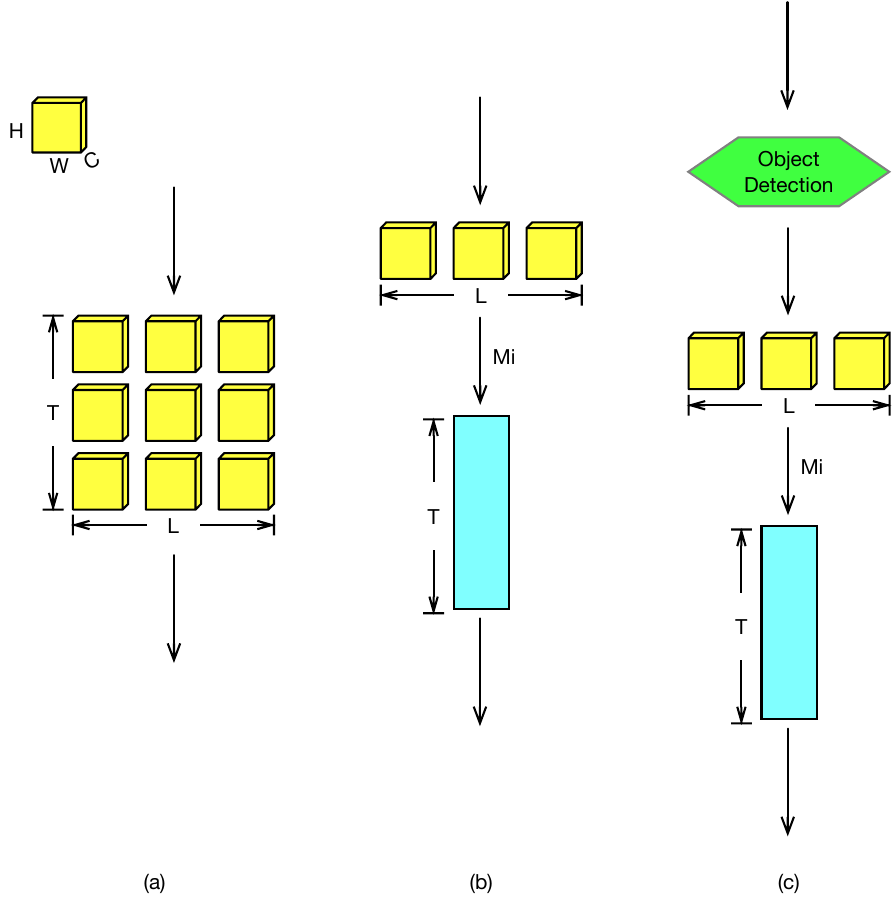}
  \caption{\textbf{5D (a) vs (4+1)D convolution (b) vs R(4+1)D (c)}. A 5D convolution employs a filter of dimensions $t \times c \times h \times h \times w \times l$. On the other hand, a (2+1)D convolutional block separates the computation into a 4D spatial convolution followed by a 1D temporal convolution. We select the $M_i$ 4D filters to ensure that the (4+1)D block corresponds to the complete 5D convolution block. For R(4+1)D, we utilize ResNet during the spatial 4D convolution phase of (4+1)D to obtain the object relationships. }
 
  \label{figure:3vs(2+1)d}
  \vspace{-0.15in}
\end{figure}
\vspace{-0.05in}
\section{Model}
\vspace{-0.05in}
In this section, we discuss the model developed in our study. Let $x$ represent the input clip of dimensions $ T \times C \times H \times W \times L$, where C refers to the number of color channels (typically 3 for RGB frames), T denotes the number of frames in the video clip, and H, W, L represent the frame height, width, and length, respectively. Given the input, we first employ the R(4+1)D model to identify the object's actions and movements. Next, we frame the problem as a VQA problem to leverage advanced NLP algorithms from ViQA research. Finally, by integrating separate models, we provide answers to open-ended relational reasoning questions from participants based on the scene and reasoning components.

\begin{figure}[ht]
  \centering
  \includegraphics[width=\linewidth,height=8cm]{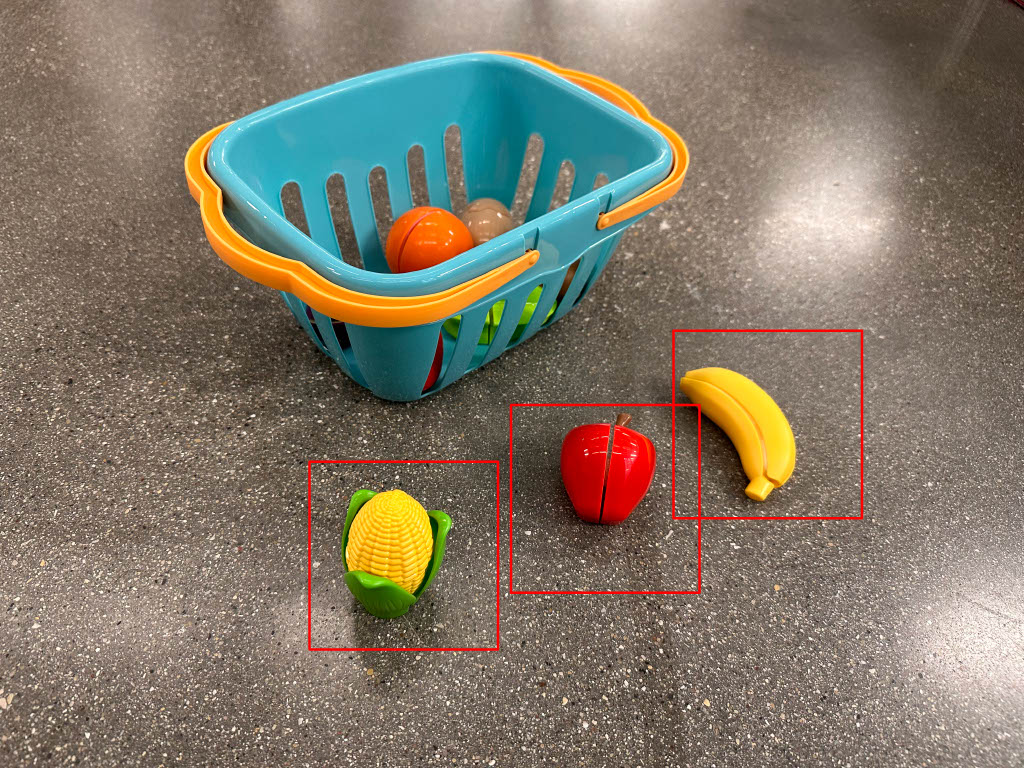}
  \caption{Examples of bounding box detection for objects actions with robot. Our robot identifies three stationary toys positioned on the ground.}
  \label{figure:bbox_robot}
  \vspace{-0.2in}
\end{figure}

Our method detects the action and motion of the object~\ref{figure:bbox_robot}.
In the previous research, many alternative technologies can complete this task with high accuracy. 
Meanwhile, in real life, there are a lot of actions that should accurately leverage temporal information. 
LSTM model \cite{greff2016lstm} is a very common solution for this purpose.
However, the LSTM lacks the experience of the object change with the rotation during the training, and it may result in a huge error in some special case. With the milestone of the I3D and I(2+1)D model, we can design a model based on those two models to detect with the frame from 1 to k and optical flow with 1 to k as well. 
However, as we mention in the related work, the I(2+1)D model is easier to optimize. 
It helps us have a high possibility to reduce the resource in the model with approximately high accuracy.  
Our model is based on the networks with 4 spatial convolutional layers and 1 temporal convolutional layer.


 In question answering, we directly apply the whisper \cite{https://doi.org/10.48550/arxiv.2212.04356} in our model to recognize the speech made by the users and transfer them to the text. We use the Stanford Parser tree and BERT \cite{devlin2018bert, https://doi.org/10.48550/arxiv.2112.02994} to encoding the text. After that, we build a model based on Relation-Aware Self-Attention \cite{wang2021ratsql} (as eq. \ref{eq:2} ). It is a model for embedding semi-structured input sequences that encodes both pre-existing relational structures and generated "soft" interactions between sequence components. Our methods for text embedding and linking are readily apparent as framework-implemented features.
 
Consider a set of inputs $\mathrm{X} = { x_i\ |\ i = 1, 2, \dots, n }$, with each $x_i \in \mathbb{R}^{d_x}$. Generally, $x_i$ is considered an unordered set; however, positional embeddings can be incorporated into $x_i$ to create an explicit order relation. A self-attention encoder, also known as Transformer, was introduced by \cite{https://doi.org/10.48550/arxiv.1706.03762} (as shown in eq \ref{eq:1}). It consists of a series of self-attention layers, where each layer transforms the input $x_i$ into $y_i \in \mathbb{R}^{d_x}$ as follows:
\begin{equation}
\begin{gathered}
    e_{ij}^{h} = \frac{x_iW_Q^h(x_jW_K^h)^T}{ \sqrt{d_x/H}} \\
    a_{ij}^h = \underset{j}{\text{softmax}}\left\{ e_{ij}^h\right\} \\
    z_i^h = \sum_{j=1}^n \alpha_{ij}^h(x_jW_V^h) \\
    z_i = \text{Concat}(z_i^1,\cdots,z_i^H) \\
    \tilde{y_i} = \text{LayerNorm}(x_i+z_i) \\
    y_i = \text{LayerNorm}(\tilde{y_i} + \text{FC}(\text{ReLU}(\text{FC}(\tilde{y_i}))))
\end{gathered}
\label{eq:1}
\end{equation}

In this context, FC denotes a fully connected layer, while LayerNorm represents layer normalization as described by the provided reference \cite{https://doi.org/10.48550/arxiv.1607.06450}. Additionally, $1 \leq h \leq H$ and $W_Q^h, W_K^h, W_V^h \in \mathbb{R}^{d_x \times (d_x/H)}$.

Based on the concept of incorporating relative position information within a self-attention layer \cite{shaw-etal-2018-self}, equation \ref{eq:1} can be modified as follows:


\begin{align}
e_{ij}^{h} &= \frac{x_iW_Q^h(x_jW_K^h + \textcolor{red}{r_{ij}^K})^T}{\sqrt{d_x/H}} \nonumber\\
z_i^h &= \sum_{j=1}^n \alpha_{ij}^h(x_jW_V^h + \textcolor{red}{r_{ij}^V})
\label{eq:2}
\end{align}

In this case, the terms $r_{ij}$ encode the known relationship between elements $x_i$ and $x_j$.


\begin{figure*}[tb]
  \centering
  {\includegraphics[height= 2.5in, width=0.45\textwidth]{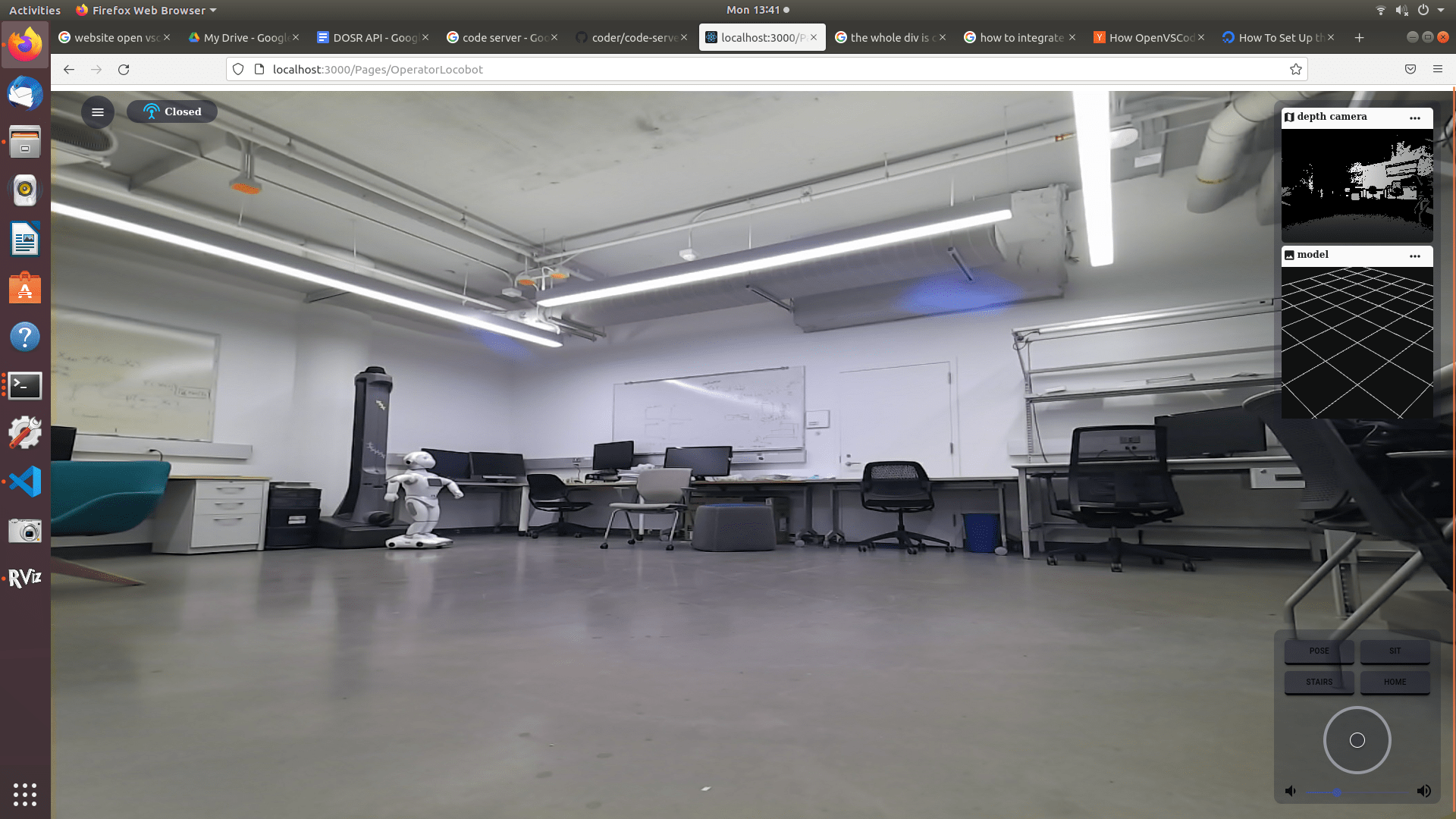}
  }
  \hfill
  {\includegraphics[height= 2.5in, width=0.45\textwidth]{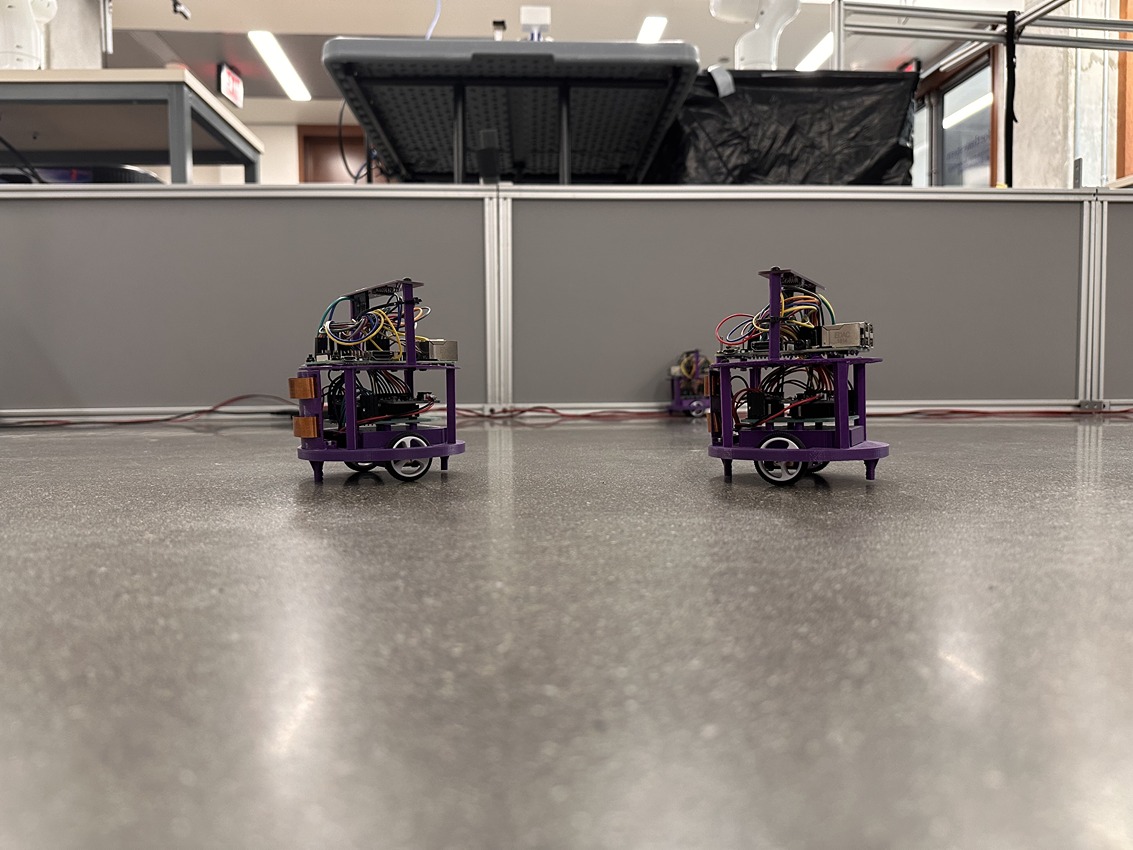}
  }

   \caption{\small \textbf{RHS: The interface of the Robot camera capture } 
  \label{fig:interface}
    \textbf{LHS: The example of robot video capturing for the movement of two small robots } }
    \label{fig:capture}
    \vspace{-0.2in}
\end{figure*}

\section{Implement}
We have created our video question answering implementation using Python3 within the ROS \cite{quigley2009ros} system. Our work utilizes the Kobuki robot, a cost-effective mobile research platform designed for education and cutting-edge robotics research. During our experiments, the robot's camera records video of the surrounding environment, while a voice recognizer installed on the Kobuki processes audio input from users. The ROS system facilitates seamless integration with our Python-based video question-answering code. Our main contribution lies not only in the integration of these components but also in providing a user-friendly API and system for practical applications.

We construct our video question-answering models using PyTorch \cite{paszke2019pytorch}, and pre-train them on an NVIDIA GeForce RTX 2080 Ti before deploying them to the Kobuki robot.

To enhance visualization, we create an interface to display video recordings captured by the Kobuki robot. Utilizing JavaScript and AJAX \cite{garrett2005ajax}, we establish a connection with the ROS system, enabling real-time visualization of the robot's camera footage. This allows users to better understand the video capture process and the robot's perspective better understand how the video captured by the robot camera work. 
\vspace{-0.05in}
\section{Experiment}
\vspace{-0.05in}
We formulate several hypotheses to support our experiment and provide information about the participants involved. Additionally, we describe the type of agent our robot is operating on. Following this, we discuss the experimental settings and the evaluation methodology employed to assess the outcomes.
\vspace{-0.05in}
\subsection{Hypothesis}
\vspace{-0.05in}
We develop four hypotheses for our experiment.
\begin{itemize}
  \item \textbf{Hypothesis 1}. The integration of individual words describing location, color, and shape enables individuals to form a fundamental understanding of the environment.
  
  \item \textbf{Hypothesis 2}. The interaction between the robot and human is designed to be straightforward, and we expect that redundant descriptions will not occur throughout the experiment. 

  \item \textbf{Hypothesis 3}. Users are receptive to the response format provided by the robot and feel comfortable when interacting with the robot system.

  \item \textbf{Hypothesis 4}. Users have confidence in the robot's responses and do not express doubt or uncertainty regarding the information provided by the robot.
\end{itemize}

\vspace{-0.05in}
\subsection{Participants}
\vspace{-0.05in}
The study involves two hundred college students as participants, all of whom are randomly selected. The two hundred participants are divided into four groups, with each group assigned to verify a specific hypothesis. Prior to the experiment, all participants are given sufficient time to familiarize themselves with their tasks and the devices. During the experiment, all participants, regardless of any pre-existing visual impairments, have their eyes covered. They are permitted to interact with the experimental device solely through speaking and listening.
\vspace{-0.05in}
\subsection{Robot Agent}
\vspace{-0.05in}
In our experiment, we establish the core functionalities within a Kobuki Robot, as depicted in Figure \ref{figure:robot}. These features include processing voice input and converting the audio clip into English, a VQA program to analyze the video and the posed question, and a text-to-speech system to vocalize the results obtained from the VQA program. To ensure consistency in the video data, the test robot maintains the same angle of view as the participants throughout the experiment, capturing footage that aligns with what the participants would have seen.

\begin{figure*}[tb]
  \centering
   
  {\includegraphics[width=0.45\textwidth,height=7cm]{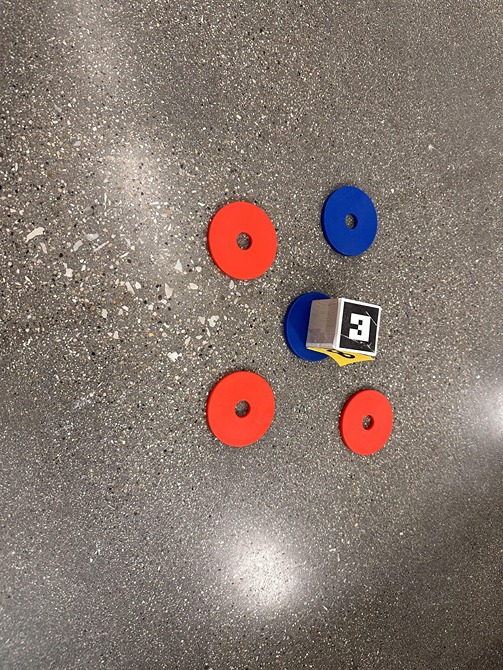}
  }
  \hfill
  {\includegraphics[width=0.45\textwidth,height=7cm]{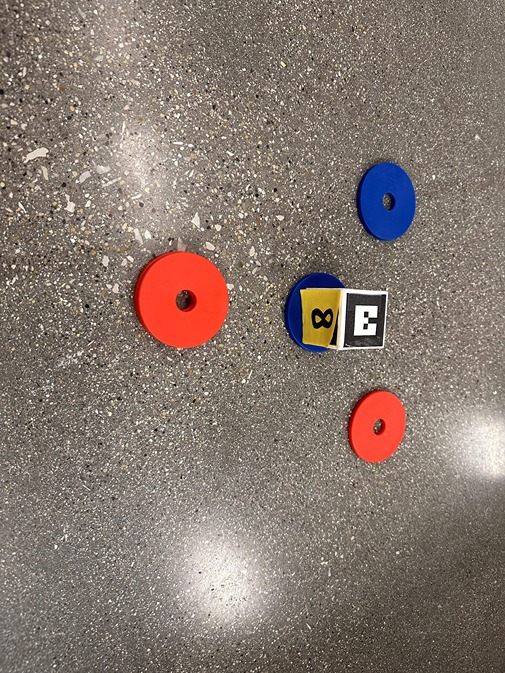}
  }

   \caption{\small \textbf{RHS: The video slides of objects before the red ring moving } 
  \label{fig:slide_before}
    \textbf{LHS: The video slides of objects after the red ring moving, the red ring moving, and put with another red ring together} }
    \label{fig:slide_after}
    \vspace{-0.2in}
\end{figure*}

\vspace{-0.05in}
\subsection{Four Experiments}
\vspace{-0.05in}
\hspace{0.5cm}\textbf{Experiment for H1:}

\hspace{0.5cm}In this experiment, participants are required to understand the environment by asking questions to the robot agent. We use video clips containing various objects, with a sample shown in Figure \ref{fig:slide_before}. The video clips in the test set offer several benefits. They display objects of different shapes and colors that frequently change positions. The relative locations can be described using simple terms, such as "near the cube" or "behind the sphere." This simplifies the real world without omitting key information (location, color, shape, action). The actions of the objects within the video are assessed based on changes in other key information.

Participants are expected to determine the answers to the following questions: How many objects are there for each distinct shape? What colors are they? What are their locations and actions? They have a maximum of thirteen minutes to complete their tasks. Following this, they are asked to create a drawing based on the information they have gathered.

All human questions and robot responses are recorded in the ROS log for later analysis.

\hspace{0.5cm}\textbf{Experiment for H2:}

\hspace{0.5cm}We analyze the log files from the previous experiments and categorize the questions into three types:
\textbf{1.} { \sf Questions that were followed by correct answers. }
\textbf{2.} {\sf Questions that were followed by incorrect answers. For example, the correct answer should be "four cubes," but the actual answer given is "one cube." }
\textbf{3.} {\sf Questions that were followed by invalid answers. For example, the correct or acceptable answer is "one sphere," but the actual answer provided is "one triangle." }
In this group of participants, some individuals only ask simple and clear questions like "How many spheres?" while others pose more complex questions, such as "Could you tell me about the environment and how many objects look like a cycle in there?" We maintain a log of these interactions and analyze the accuracy of the responses provided by the robot agent.

\hspace{0.5cm}\textbf{Experiment for H3:}

\hspace{0.5cm}In this experiment, participants are free to ask questions with the objective of understanding the environment. Two different robot agents are used for the participants. One robot agent is programmed to be polite and helpful, while the other is programmed to be uncooperative.
The polite robot initiates conversations with friendly phrases such as "How may I assist you?" or "Is there anything I can do for you?" If this robot is unable to process a question, it might respond with "Sorry, this little robot cannot find the answer. Would you mind rephrasing your question in a simpler way?" In contrast, the uncooperative robot provides minimal responses such as "one" or "None."
After the question-asking phase, participants are asked to create a drawing based on the information they have gathered from their interactions with the robot agents.

\hspace{0.5cm}\textbf{Experiment for H4:}

\hspace{0.5cm}Among the participants, they complete a survey that includes questions about their trust. We evaluate the survey results and identify one participant who trusts the robot the most and another participant who trusts the robot the least. Both participants interact with the "friendly" robot agent and subsequently create drawings based on the information they have gathered during their interactions.
\vspace{-0.05in}
\subsection{Method of Evaluation}
\vspace{-0.05in}
We utilize accuracy as our evaluation metric. The accuracy can be calculated with:\newline
$$Accuracy = \frac{Correct \ answer \ number }{Total \ answer \ of \ that \ problem}$$

For each type of object (Cube, Sphere, Cone), they are present in different quantities. We randomly select videos from the CATER dataset and ask the participants to inquire about the key information (location, color, shape) of the objects. The participants are then required to draw these objects based on the information they have gathered. For each object type, we assess whether their answers match the ground truth and calculate the accuracy score.

Lastly, we evaluate the participants' understanding of the relative locations of the objects. We ask them to pose questions about the locations of each object at five random time points. The participants are then required to draw the spatial relationships between the objects at each of these time points. As with the previous assessments, we compare their answers to the ground truth and calculate the accuracy score.

\vspace{-0.05in}
\section{Result And Discussion}
\vspace{-0.05in}
We conclude that our methods not only provide reliable numerical results, but they are also user-friendly and foster effective human-robot interaction.
\vspace{-0.05in}
\subsection{Performance of ViQA Model}
\vspace{-0.05in}

\begin{table}[ht]
\caption{Performance Comparison between Models} 
\centering 
\begin{tabular}{lllll} 
\toprule 
Method & Count$\%$ & Color $\%$ & Shape $\%$ & Location $\%$\\ [0.5ex]  
\midrule 
Lei et al. ~\cite{lei2018tvqa}& 69.32 & 68.23 & 64.34 & 53.72 \\ 
 Jie et al. ~\cite{kim2019progressive} & 70.16 & 73.32 & 69.81  & 57.34 \\
 Jie et al. ~\cite{kim2019gaining} & 72.81 & 73.23 & 71.45 & 61.23\\
 Yang et al. ~\cite{yang2020bert} & 75.13 & \textbf{74.97} & \textbf{77.14} & 63.76\\
 Chadha et al. ~\cite{chadha2020iperceive} & 77.23 & 74.31 & 74.56 & 62.56\\
 R(2+1)D & \textbf{80.42} & 72.67 & 75.23 & \textbf{65.71}\\ [1ex] 
\bottomrule 
\end{tabular}
\label{table:table1} 
\end{table}

We have compared the classification accuracy of our model with several other models (Table ~\ref{table:table1}) in experiments 1 and 2. For methods used in CATER and Chadha et al. ~\cite{chadha2020iperceive}, they employ attention algorithms to ensure the model is competitive for the task. However, they did not perform as well as our model. All models and their variables were trained once per participant in experiment one, resulting in an average performance. We find that these previous algorithms did not perform as well as our model. Our model outperforms Chadha et al. ~\cite{chadha2020iperceive} by 3.19

The improvements offered by our model hold significant implications for other models. However, the performance of our model in some features of the ViQA task is not as good as that of other models. For instance, our model's accuracy in judging an object's color is lower than that of Chadha et al. ~\cite{chadha2020iperceive}. The reason  is that our model focuses more on the relationship between object movements. In our model, we use optical flow to measure an object's actions. When employing optical flow, we do not use RGB3, causing our model to lose color information in some situations and resulting in a higher error rate in model performance.

\subsection{Trust and Interactivity of HRI}

\begin{figure}[ht]
\vspace{-0.05in}
  \centering
  \includegraphics[width=\linewidth]{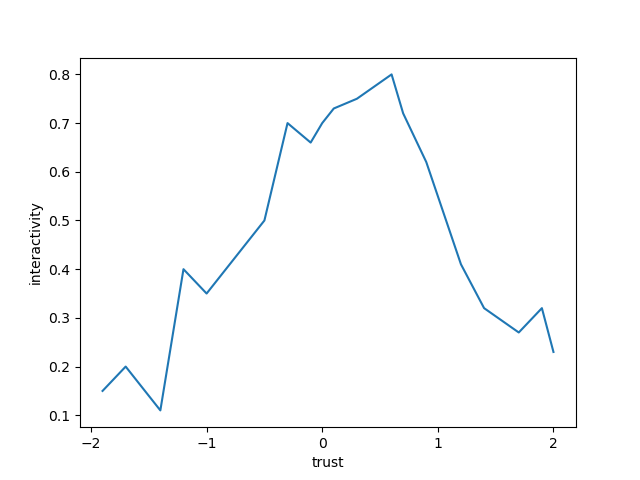}
  \vspace{-0.2in}
  \caption{Diagram about trust and interactivity. Individuals who perceive a higher degree of interactivity in a robot are more inclined to trust the system. To facilitate comparison and analysis, we normalize the interactivity score on a scale of 0-1, derived from the original 0-10 scale. Similarly, the trust score is normalized from the 0-10 scale to a -2 to 2 scale.}
  \label{figure:fig5}
   \vspace{-0.1in}
\end{figure}

\begin{figure}[ht]
  \centering
  \vspace{-0.05in}
  \includegraphics[width=\linewidth]{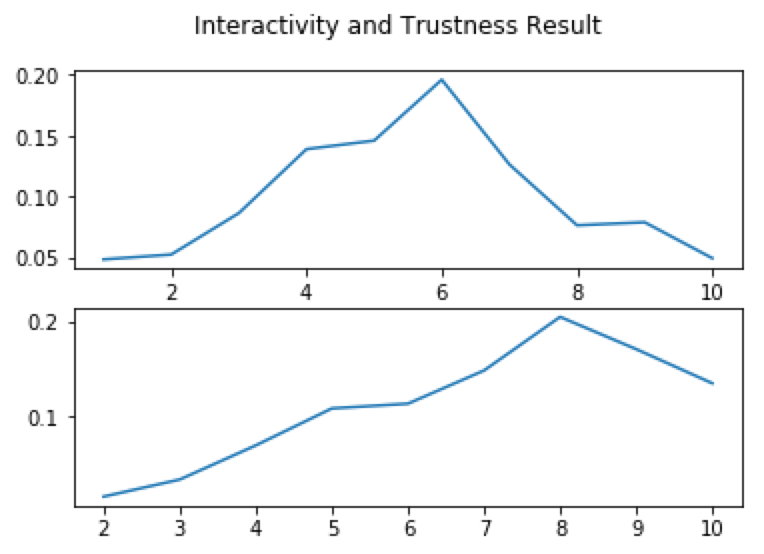}
   \vspace{-0.2in}
  \caption{Diagram for trust and interactivity individually. The two-part diagram illustrates interactivity and trust levels in our human-robot interaction system. The upper diagram displays the interactivity level, while the lower diagram showcases the trust level. On the y-axis, the percentage of participation is represented, while the x-axis denotes user ratings, ranging from 0 to 10. Based on user feedback, it has been observed that our system exhibits both high interactivity and trust levels for users. }
  \label{figure:fig6}
  \vspace{-0.15in}
  
\end{figure}

We use a survey to calculate the trust score for each participant. The lower the trust score, the less trust the participant has in the robot. A trust score between $-1$ and $+1$ is a reasonable range, indicating that participants neither overtrust nor undertrust the robot. Participants with low trust in the robot are less likely to have efficient interactions with it. For instance, a participant who asked the robot to identify object locations and relied on its guidance but distrusted the robot might take longer to visualize the surroundings or reach their desired destination. It is essential to understand why undertrust leads to poorer interactions. Ideally, a participant receiving the robot's feedback should make a bold attempt based on the provided information, while also considering whether the feedback is reasonable.

Figure ~\ref{figure:fig6} represent the interactivity and trust results in a plot diagram and histogram diagram. In Figure ~\ref{figure:fig6}, for example, the upper diagram represents interactivity, and the lower diagram represents trust. On the x-axis, we scale from 1 to 10, representing the scores we used to evaluate the participants. The y-axis represents the probability of the scores occurring in our survey.

From Figure ~\ref{figure:fig6} we observe a positive correlation between trust and interactivity. Consequently, we project the experimental results of the 200 participants to produce Figure ~\ref{figure:fig5}.

\vspace{-0.05in}
\section{Conclusion And Future Work}
\vspace{-0.05in}v
We developed a new model that facilitates high-level interactions with blind users, focusing primarily on interaction improvement. The model employs VQA techniques and exhibits varying performance in different scenarios. To determine the optimal settings for the best interaction results, we designed the first two hypotheses. We also compared our model with other models in the ViQA domain, finding that our model outperforms others in most areas. Furthermore, we investigated the effects of trust on interaction and identified features that positively impact the interaction between our robot agent and blind individuals.

Our current experiments utilize "object video." In future work, we plan to deploy our robot agent in more complex and realistic situations. We may also equip our robot agent with mechanical arms, enabling it to perform actions beyond linguistic communication. This modification will alter the interaction between the new robot agent and blind people. We intend to examine the limits of the new robot agent's capabilities and explore potential improvements. Additionally, we plan to incorporate different types of attention mechanisms into our model, such as hierarchical attention ~\cite{yang2016hierarchical}. We will also strive to enhance our model's color detection sensitivity by integrating optical flow with RGB3.


\bibliographystyle{IEEEtran}
\bibliography{base}

\end{document}